\documentclass[10pt,twocolumn,letterpaper]{article}

\usepackage{iccv}
\usepackage{times}
\usepackage{epsfig}
\usepackage{graphicx}
\usepackage{amsmath}
\usepackage{amssymb}
\usepackage{listings}
\usepackage[pagebackref=true,breaklinks=true,letterpaper=true,colorlinks,bookmarks=false]{hyperref}
\usepackage[capitalize]{cleveref}
\usepackage{booktabs}
\usepackage{enumitem}
\usepackage{caption}
\usepackage{xcolor}
\definecolor{Blue}{RGB}{0,90,255}
\definecolor{Red}{RGB}{255,75,0}
\definecolor{Green}{RGB}{3,175,122}
\hypersetup{
colorlinks=true,bookmarksopen=true,citecolor=Blue,urlcolor=Red,linkcolor=Green
}

\usepackage[breaklinks=true,bookmarks=false]{hyperref}
\usepackage[accsupp]{axessibility}

\iccvfinalcopy 

\def\iccvPaperID{9406} 

\ificcvfinal\fi

\begin{document}

\title{Will Large-scale Generative Models Corrupt Future Datasets?}

\author{Ryuichiro Hataya\\
RIKEN ADSP \& RIKEN AIP\\
{\tt\small ryuichiro.hataya@riken.jp}
\and
Han Bao\\
Kyoto University\\
{\tt\small bao@i.kyoto-u.ac.jp}
\and
Hiromi Arai\\
RIKEN AIP\\
{\tt\small hiromi.arai@riken.jp}
}

\maketitle
\ificcvfinal\thispagestyle{empty}\fi

\begin{abstract}
Recently proposed large-scale text-to-image generative models such as DALL$\cdot$E 2~\cite{dalle2}, Midjourney~\cite{midjourney}, and StableDiffusion~\cite{stablediffusion} can generate high-quality and realistic images from users' prompts. 
Not limited to the research community, ordinary Internet users enjoy these generative models, and consequently, a tremendous amount of generated images have been shared on the Internet.
Meanwhile, today's success of deep learning in the computer vision field owes a lot to images collected from the Internet.
These trends lead us to a research question: ``\textbf{will such generated images impact the quality of future datasets and the performance of computer vision models positively or negatively?}''
This paper empirically answers this question by simulating contamination. Namely, we generate ImageNet-scale and COCO-scale datasets using a state-of-the-art generative model and evaluate models trained with ``contaminated'' datasets on various tasks, including image classification and image generation.
Throughout experiments, we conclude that generated images negatively affect downstream performance, while the significance depends on tasks and the amount of generated images.
The generated datasets and the codes for experiments will be publicly released for future research.
Generated datasets and source codes are available from \url{https://github.com/moskomule/dataset-contamination}.
\end{abstract}

\section{Introduction}\label{sec:introduction}

\begin{figure}[t]
	\centering
	\includegraphics[width=0.5\linewidth]{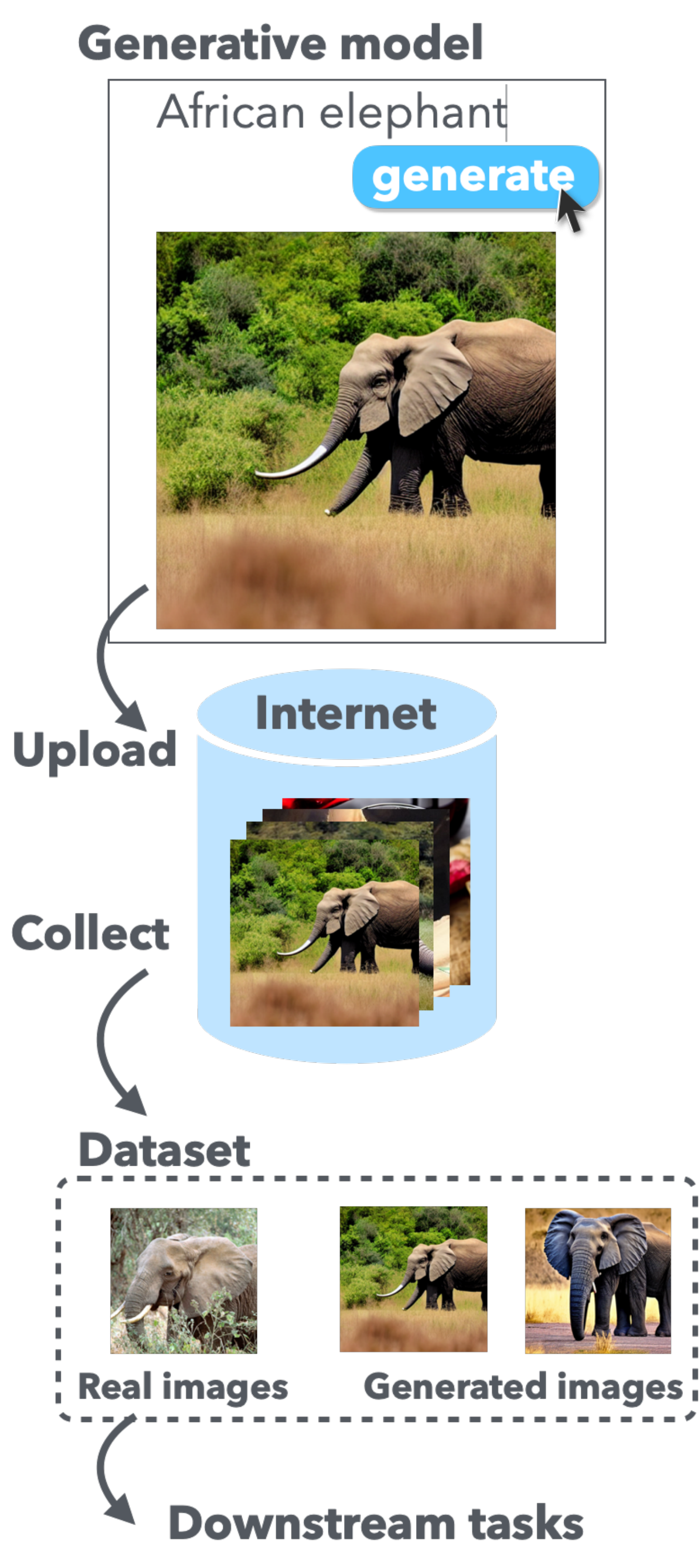}
	\caption{Schematic view of the problem. Some large-scale generative models are public, and many users are playing with them to share generated images on the Internet (top). Dataset collection heavily relies on images on the Internet, which may be contaminated by generated images (bottom). This paper discusses the effect of such dataset corruption.}
	\label{fig:schematic}
\end{figure}

Deep generative models for image generation have advanced progressively since the original GANs~\cite{Goodfellow2014b} and VAEs~\cite{kingma2014vae}.
Recently, denoising diffusion models~\cite{ho2020denoising,sohl2015deep,song2021scorebased} have beaten GANs in image quality~\cite{dhariwal2021,ho2022cascaded} and become one of the de-facto standard generative models.
Among them, some large models trained with billion-scale captioned images collected from the Internet achieved high-fidelity image generation conditioned by users' prompts~\cite{dalle2,saharia2022photorealistic,stablediffusion,midjourney,nichol22a,zhang2021ernie,yu2022scaling,gafni2022make,balaji2022ediffi}.
Particularly, DALL$\cdot$E 2~\cite{dalle2}, Midjourney~\cite{midjourney}, and StableDiffusion~\cite{stablediffusion} have web and smartphone applications, and many Internet users enjoy image generation, and consequently, a tremendous amount of generated images have been uploaded to the Internet.\footnote{The DALL$\cdot$E 2 model alone generated two million images per day in September 2022, according to \url{https://openai.com/blog/dall-e-now-available-without-waitlist/}.}

At the same time, highly realistic generated images may have potentially significant impacts on society.
For example, face images generated by StyleGANs~\cite{karras2019style} were reportedly used to create fake profiles of SNSs or dating apps to deceive other users~\cite{harwell2020dating,hill2020designed}.
Furthermore, recent text-to-image generative models can generate images that look real at first glance from users' instruction and are able to support fake news~\cite{tiku2022}.
They also amplify demographic stereotypes~\cite{federico2022}.

Another concern is that generated images might affect the quality of newly curated image datasets from the Internet in the future, similar to the fact that the outputs of machine translation models degenerate the quality of corpora~\cite{simard2014clean,rarrick2011mt,dodge2021}.
Without a doubt, today's success of deep learning and computer vision, including generative models themselves, largely owes to image datasets collected from the Internet, such as ImageNet~\cite{Russakovsky2015,deng2009imagenet}.
However, when generated images are shared on the Internet, they may contaminate the sources of image datasets (see also \cref{fig:schematic}).\footnote{Although the ``official'' web applications implant watermarks to generated images, which thus can be filtered, we found that some software has options to disable such functions, \eg, \url{https://github.com/AUTOMATIC1111/stable-diffusion-webui}.}
Based on these backgrounds, our research question in this paper raises: \emph{what will happen if datasets were contaminated by generated images?}

We aim to answer this question through experiments: simulating such contamination by large-scale datasets of generated images and measuring downstream performance trained with them.
Specifically, we generate three million images from ImageNet categories and COCO captions using StableDiffusion and emulate the contamination by replacing real images in datasets with generated ones. 
Then, we measure the performance of models trained with such contaminated datasets in various tasks, namely, image classification, image captioning, and image generation.
Throughout experiments, we find that generated images have \emph{negative} effects on downstream performance.
We hypothesize that such negative effects are caused by the fact that the generative models capture fewer modes than the actual data, although the existing synthetic experiments have shown high coverage~\cite{xiao2022DDGAN}.

In summary, our contributions are as follows:

\begin{itemize}[itemsep=0pt,topsep=0pt,parsep=0pt,leftmargin=*]
    \item To simulate the effects of possible contamination, we create large-scale datasets consisting of generated images corresponding to ImageNet and COCO caption (\cref{sec:method}).
    \item We conduct experiments over four distinct tasks on the generated datasets and discover negative effects of contamination, which can be partially attributed to fewer modes of generated images than real data (\cref{sec:experiments,sec:analysis}). 
    \item Based on the empirical results, we recommend to researchers how to publish generative models and how to collect datasets (\cref{sec:conclusion}).
\end{itemize}

\section{Background and Related Work}

A deep generative model aims to approximate the underlying data distribution by neural networks,  and the sampled data are expected to be similar to real ones.
Since the emergence of GANs~\cite{Goodfellow2014b,radford2015unsupervised}, the research of deep generative models has advanced progressively.
In particular, denoising diffusion models, equivalently, score-based generative models, have achieved high-quality image generation with diversity~\cite{ho2020denoising,sohl2015deep,song2021scorebased}, capturing modes of data distributions faithfully~\cite{xiao2022DDGAN}.
Text-to-image generative models based on diffusion models can generate high-quality images from users' text instructions with high fidelity, even for unseen novel combinations of concepts, such as ``a photo of an astronaut riding a horse''~\cite{dalle2,saharia2022photorealistic,stablediffusion,midjourney,nichol22a,zhang2021ernie,yu2022scaling,gafni2022make,balaji2022ediffi}. 
Notably, some models have publicly accessible applications~\cite{stablediffusion,midjourney,dalle2,zhang2021ernie}, and many Internet users are generating images and posting them on the Internet with related texts.
Such generated images are sometimes difficult to be distinguished from real ones, and thus some of them are potent to contaminate future datasets collected from the web.

The NLP community has experienced similar problems in the last decade; thanks to the development of NLP technologies, many contents on the Internet have become machine-generated, \eg, by machine translation and optical character recognition systems, but such generated texts have degenerated quality of corpora~\cite{simard2014clean,dodge2021}.
As a result, filtering such low-quality samples is essential to maintain downstream performance~\cite{rarrick2011mt}.
Although the NLP community has investigated such issues by generated data, the effects of generated images by text-to-image models on various downstream performances in computer vision have rarely been studied.

The dataset contamination issue in general has been studied from various aspects; including dataset poisoning~\cite{chen2017targeted}, adversarial training~\cite{goodfellow2014explaining}, label noise~\cite{han2020survey}, outlier robustness~\cite{huber2011robust}, and distribution shifts~\cite{sinha2018certifiable}. The existing studies usually posit an attacker/contamination model that is plausible yet mathematically convenient, such as Huber’s contamination model~\cite{huber2011robust}.
By contrast, we are rather interested in realistic contamination of the web images induced by generative models and its potential effects.

Methodologically, our work is related to~\cite{ravuri2019cl}; they also use accuracy on ImageNet classification, but their purpose is to measure how much a generative model captures the ImageNet data distribution.
Differently, our interest is not in evaluation of generative models, but the effect of generated images on various downstream tasks, and the image classification task is used along with others.

\section{Dataset Corruption by Generated Images}\label{sec:method}

This paper aims to answer our research question ``\emph{will the contamination of generated images perform positively or negatively?}''
To empirically answer this question, we simulate realistic dataset contamination by generated images and evaluate the quality of contaminated datasets by training commonly-used models with such datasets in several tasks. In this section, we describe the dataset creation.

\subsection{Dataset Creation}

To simulate image generation by users, we create datasets using a StableDiffusion model~\cite{stablediffusion}, a state-of-the-art text-to-image generative model, pre-trained with LAION-2B~\cite{schuhmann2022laionb}, which include not all but at least some ImageNet and COCO images.
These datasets are generated from category names of the ImageNet ILSVRC-2012 classification dataset and captions of the COCO caption dataset, which are referred to as SD-ImageNet and SD-COCO in the remaining text.
For generation of both datasets, we disabled the watermarking functionality to trace outputs as generated images and the safety checker to reduce explicit outputs, such as nudes.

\begin{figure*}[t]
	\centering
	\includegraphics[width=\linewidth]{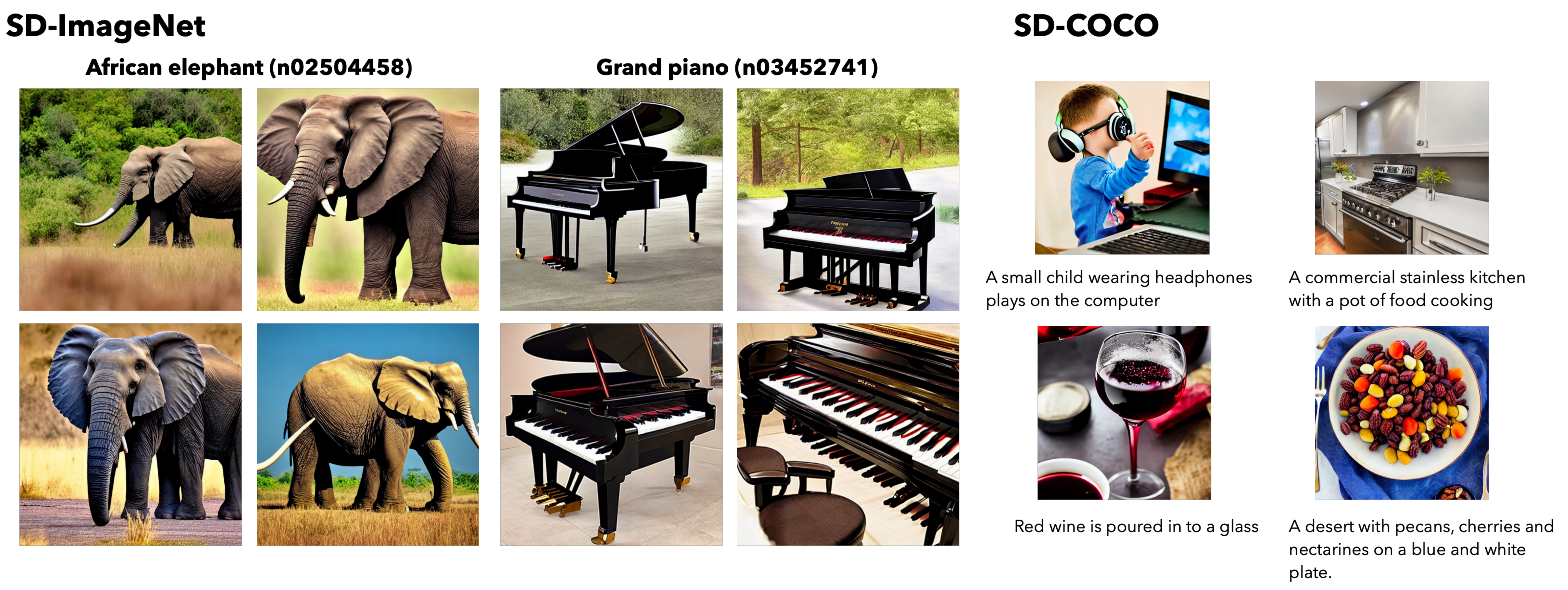}
	\caption{Randomly selected examples from generated datasets, namely, SD-ImageNet (left) and SD-COCO (right). Images are at a single glance high quality and fidelity to prompts, \ie, category names and captions, while details are unnatural, \eg, a two-nose elephant.}
	\label{fig:datasets}
\end{figure*}

\begin{description}[itemsep=0pt,topsep=0pt,leftmargin=*]
    \item[SD-ImageNet:]
The ImageNet ILSVRC-2012 classification dataset~\cite{Russakovsky2015} is a subset of ImageNet~\cite{deng2009imagenet} and a dataset for the image classification task. Its training set contains 1.2 million photo images over 1,000 categories selected from synsets of WordNet, \eg, ``African elephant''. 
Using these category names, we prepared prompts like ``A photo of African elephant'' for each category and generated 1,400 photography-like images for each class.

\Cref{fig:datasets} (left) shows examples from SD-ImageNet. Images are natural at first glance, but contain some flaws. For example, the elephant at the top left has two noses.
We will revisit the creation of prompts in \cref{sec:prompts}.

\item[SD-COCO:]
The COCO caption dataset~\cite{chen2015microsoft} is a dataset for the image captioning task. 
Based on the dataset split in~\cite{karpathy2015deep}, this dataset has 113,000 images with five captions for each image, such as ``A small child wearing headphones plays on the computer''. These captions were used as prompts to generate 565,000 images.

\Cref{fig:datasets} (right) presents some examples from SD-COCO with their captions. Similar to examples of SD-ImageNet, the images are apparently faithful to the captions used as prompts, but staring at them reveals unnatural or unfaithful details. 
For example, the bottom right example fails to produce a ``blue and white plate.''

In the remaining text, we call the ILSVRC-2012 dataset as ImageNet and the COCO caption dataset as COCO for simplicity.
    
\end{description}

\subsection{Simulation of Corruption}\label{sec:dataset_generation}

To simulate possible corruption, we randomly substitute generated images for 20, 40, and 80 \% of real images of the original datasets with generated ones without replacement.
We refer to these mixed datasets as IN/SD-$n$\%, where $n$ indicates the ratio of generated data.
Similarly to IN/SD-$n$\%, we also created mixtures of COCO and SD-COCO, which are referred to as CO/SD-$n$\%.

In the next section (\cref{sec:experiments}), we empirically investigate the effect of corruption using the downstream performance of models trained with these contaminated datasets.
However, mixtures of ImageNet and SD-ImageNet, \eg, IN/SD-20\%, alone still entangle the effect of artifacts of generated images and the domain shift between generated images and real images.
To disentangle the effect of generated images from that of domain shift, we additionally use datasets consisting of real images similar to ImageNet and COCO and compare the downstream performance of models trained with them with that of IN/SDs or IN/COs.
As a counterpart of ImageNet, we adopt a subset of the WebVision dataset~\cite{li2017webvision}, which was collected by querying ImageNet category names to Google and Flickr; then, we mix it with ImageNet. 
Because this subset is imbalanced, and some of its categories contain fewer images than needed, we then use sampling by replacement to create balanced mixtures. Similar to IN/SD-$n$\%, we refer to these mixed datasets as IN/WV-$n$\%.
Correspondingly, as a counterpart of COCO, we use Flickr-30k~\cite{flick30kb}, which contains 32,000  images collected from Flickr with five captions per image.
Because its size is much less than COCO, we only prepare CO/FL-40\% as a mixture of COCO and Flickr-30k.

\section{Experimental Results}\label{sec:experiments}

In this section, we evaluate the effect of contamination using the datasets created in \cref{sec:method} on several downstream tasks.

\subsection*{Shared Experimental Settings}

We used neural network models implemented with \texttt{PyTorch} v1.12~\cite{pytorch} and its accompanying \texttt{torchvision} with CUDA 11.3. 
Experiments including dataset creation described in \cref{sec:method} were conducted on NVIDIA V-100 GPUs and NVIDIA A-100 GPUs. Further description of settings and configurations can be found in \cref{app:config} in the Supplementary Material.

\subsection{Image Classification}

This task classifies images into 1,000 categories of ImageNet.
We used ResNet-50 \cite{He2016b}, SwinTransformer-S (Swin-S) \cite{liu2021swin}, and ConvNeXt-T \cite{liu2022} in \texttt{torchvision}, training them according to the standardized training protocols\footnote{\url{https://github.com/pytorch/vision/tree/v0.13.1/references/classification}. Exceptionally, ConvNeXt was trained for 300 epochs due to computational resources.} with ImageNet, SD-ImageNet, WebVision, and their mixtures.
ResNet is a convolutional neural network with residual connections, SwinTransformer is a variant of Vision Transformer, and ConvNeXt is a CNN inspired by Vision Transformers, which represent modern vision models.

\Cref{tab:classification} shows accuracy on the ImageNet validation set.
As can be seen, the performance decreases as the ratio of SD-ImageNet in training data increases.
When the ratio of generated images is at most 40\%, the performance drops are marginal and may be endurable in most practical scenarios.
However, when the ratio is 80\%, the performance degeneration is not negligible.
Compared to SD-ImageNet, WebVision images have less influence on performance.
In the extreme cases, when no ImageNet data are included in training data, \ie, SD-ImageNet and WebVision, this difference in performance is significant, which suggests that the performance drop may not be solely due to the domain gap.

\begin{figure}
    \centering
    \includegraphics[width=0.9\linewidth]{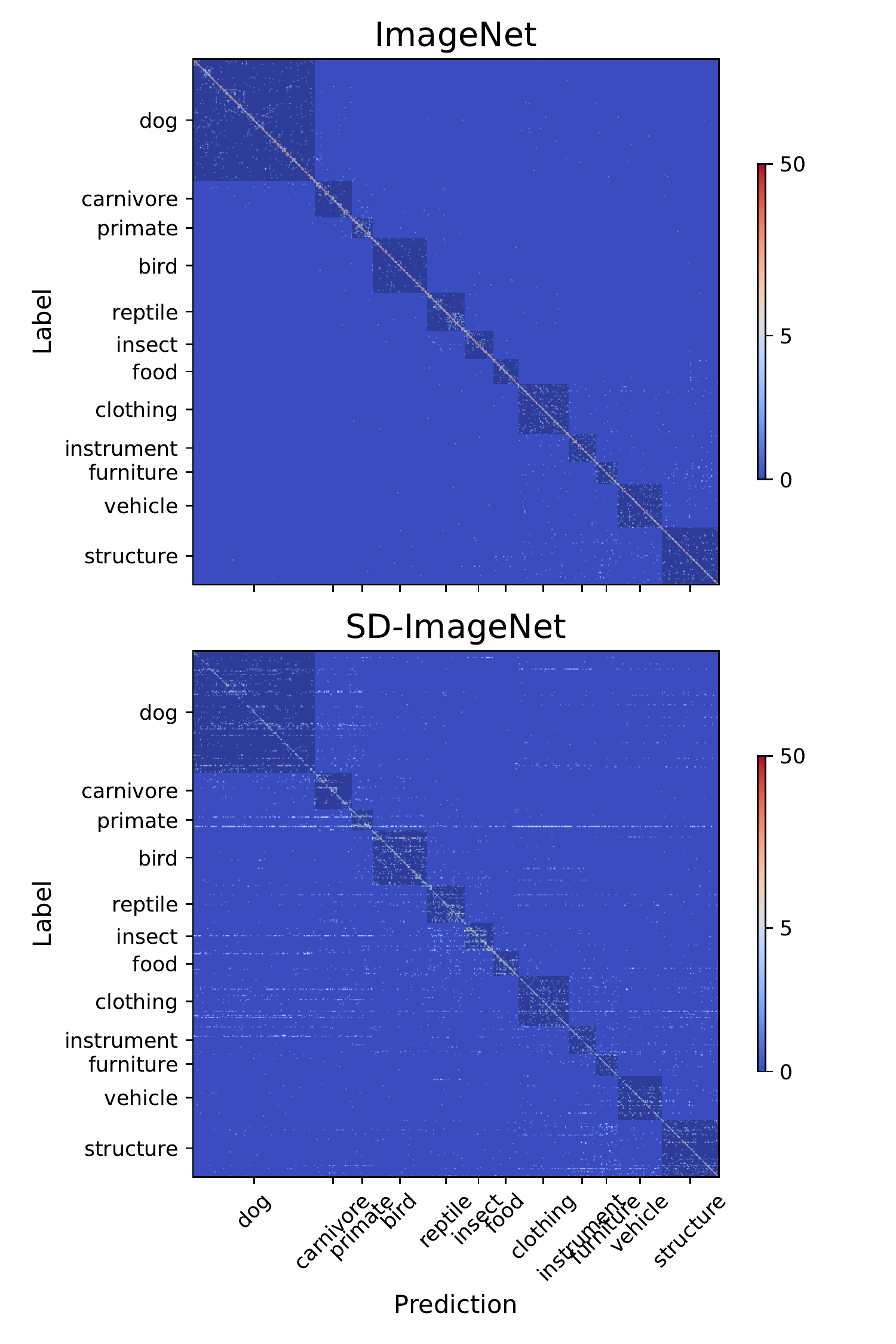}
    \caption{Confusion matrices of ResNet-50 predictions with a subset of ImageNet validation data. Models were trained with ImageNet and SD-ImageNet. Class indices are rearranged. X and Y axes depict superclasses by gathering categories. Colors correspond to the number of data at each pixel in log scale, and block-diagonal components are highlighted. Notice that the model trained with SD-ImageNet misclassifies certain classes uniformly, as illustrated by white horizontal lines.}
    \label{fig:confusion_matrix}
\end{figure}

Additionally, \cref{fig:confusion_matrix} presents confusion matrices of ResNet-50 trained with ImageNet and SD-ImageNet.
For clarity, categories are subsampled and rearranged according to 12 superclasses, adopting \texttt{big\_12} classes from~\cite{robustness}.
As can be seen, the mispredictions by ResNet-50 trained with ImageNet mostly fall in the same fine-grained categories, represented by diagonal blocks.
Contrarily, we observe that the model trained with SD-ImageNet uniformly misclassifies certain classes, partially because the category names of such classes are ambiguous, and thus, the generated images for such classes are semantically diverse.
\texttt{Titi} (monkey) is such an example, where it is intended to mean a New World monkey in ImageNet, but it is also a name of people, plants, and places, and thus, generated images are also semantically diverse (see \cref{fig:titi} in the Supplementary Material).

\begin{table}[h]
	\centering
    \resizebox{0.9\linewidth}{!}{%
	\begin{tabular}{lccc}
	\toprule
	                     & ResNet-50      & Swin-S   & ConvNeXt-T   \\
	\midrule
	ImageNet             & 75.7           & 83.1     & 80.8      \\
	\midrule
	IN/SD-20\%           & 74.5           & 82.1     & 79.7         \\
	IN/SD-40\%           & 72.6           & 81.0     & 78.3     \\
	IN/SD-80\%           & 65.3           & 74.3     & 70.8     \\
	SD-ImageNet          & 15.7           & 19.3     & 19.6      \\
	\midrule
	IN/WV-20\%           & 75.1           & 82.5     & 80.0     \\
	IN/WV-40\%           & 73.9           & 81.8     & 78.8     \\
	IN/WV-80\%           & 68.3           & NaN      & 73.9     \\
	WebVision            & 61.3           & 70.9     & 66.2      \\
	\bottomrule
	\end{tabular}
    }
	\caption{Validation accuracy of the image classification task on the ImageNet validation set. We could not stably train Swin-S on IN/WV-80, which resulted in a loss explosion. The performance drop is marginal when the ImageNet images dominate the dataset.}
	\label{tab:classification}
\end{table}

\subsection{Image Captioning}

Image captioning is a task to generate appropriate captions for given images.
We used a pre-trained BLIP model~\cite{li2022blip}, a state-of-the-art vision-language model, and fine-tuned its captioner and filter modules on COCO, SD-COCO, Flickr-30k, and their mixtures for five epochs, following~\cite{lavis}.

\Cref{tab:imagecaptioning} reports the performance in various statistics with the COCO test set when the captions are generated with beam search with a beam size of three.
Aligned with the results of image classification, a performance drop by generated images can also be observed.
Especially, CO/SD-20\% yields comparable or inferior performance to CO/FL-40\%, even in metrics for image captioning like SPICE~\cite{spice} and CIDEr~\cite{cider}, indicating that generated images cause degeneration of dataset quality.
Moreover, a comparison between the results of SD-COCO and Flickr-30k suggest that such performance drops cannot be fully attributed to domain shift.

\begin{table*}[t]
	\centering
\resizebox{\linewidth}{!}{%
	\begin{tabular}{lcccccccc}
\toprule
			&	BLEU-1	&	BLEU-2	&	BLEU-3	&	BLEU-4~\cite{bleu} &	 SPICE~\cite{spice} & METEOR~\cite{meteor}	&	ROUGE-L~\cite{rouge}	&	CIDEr~\cite{cider}	\\
\midrule 
COCO	    &	0.791	&	0.641	&	0.508	&	0.400	&	0.240	&	0.310	&	0.602	&	1.335	\\
\midrule
CO/SD-20\%	&	0.787	&	0.634	&	0.500	&	0.391	&	0.235	&	0.306	&	0.596	&	1.320	\\
CO/SD-40\%	&	0.786	&	0.632	&	0.499	&	0.390	&	0.236	&	0.307	&	0.596	&	1.319	\\
CO/SD-80\%	&	0.780	&	0.623	&	0.486	&	0.377	&	0.233	&	0.300	&	0.588	&	1.279	\\
SD-COCO   	&	0.711	&	0.534	&	0.394	&	0.287	&	0.191	&	0.252	&	0.514	&	1.000	\\
\midrule
CO/FL-40\%   &   0.787   &  0.634    & 0.501      & 0.393     & 0.238      & 0.308    &  0.598   &  1.326  \\
Flickr 30k  &   0.754   &   0.587   &   0.439   &   0.321   &   0.215   &   0.275   &   0.554   &    1.092   \\
\midrule
w/o fine-tuning & 0.473  &   0.392   &   0.308   &   0.237   &   0.158   &   0.212   &   0.488   &   0.838   \\
\bottomrule
	\end{tabular}
}
	\caption{Test metrics in image captioning of the BLIP model on the COCO test split. Higher values are better.}
	\label{tab:imagecaptioning}

\end{table*}

\subsection{Image Generation}

Finally, we verify if generated images are useful as training data for the image generation task.
We evaluated an improved denoising diffusion probabilistic model (IDDPM)~\cite{nichol21a} with datasets resized to $64\times 64$, which we refer to as, for example, ImageNet-64 and IN/SD-20\%-64. 
We trained the model for $1.8\times 10^6$ iterations using the $L_\text{hybrid}$ objective~\cite{nichol21a} with a batch size of 512 and generated $5.0\times 10^4$ images with 250 sampling steps.

\begin{figure}[t]
    \centering
    \includegraphics[width=\linewidth]{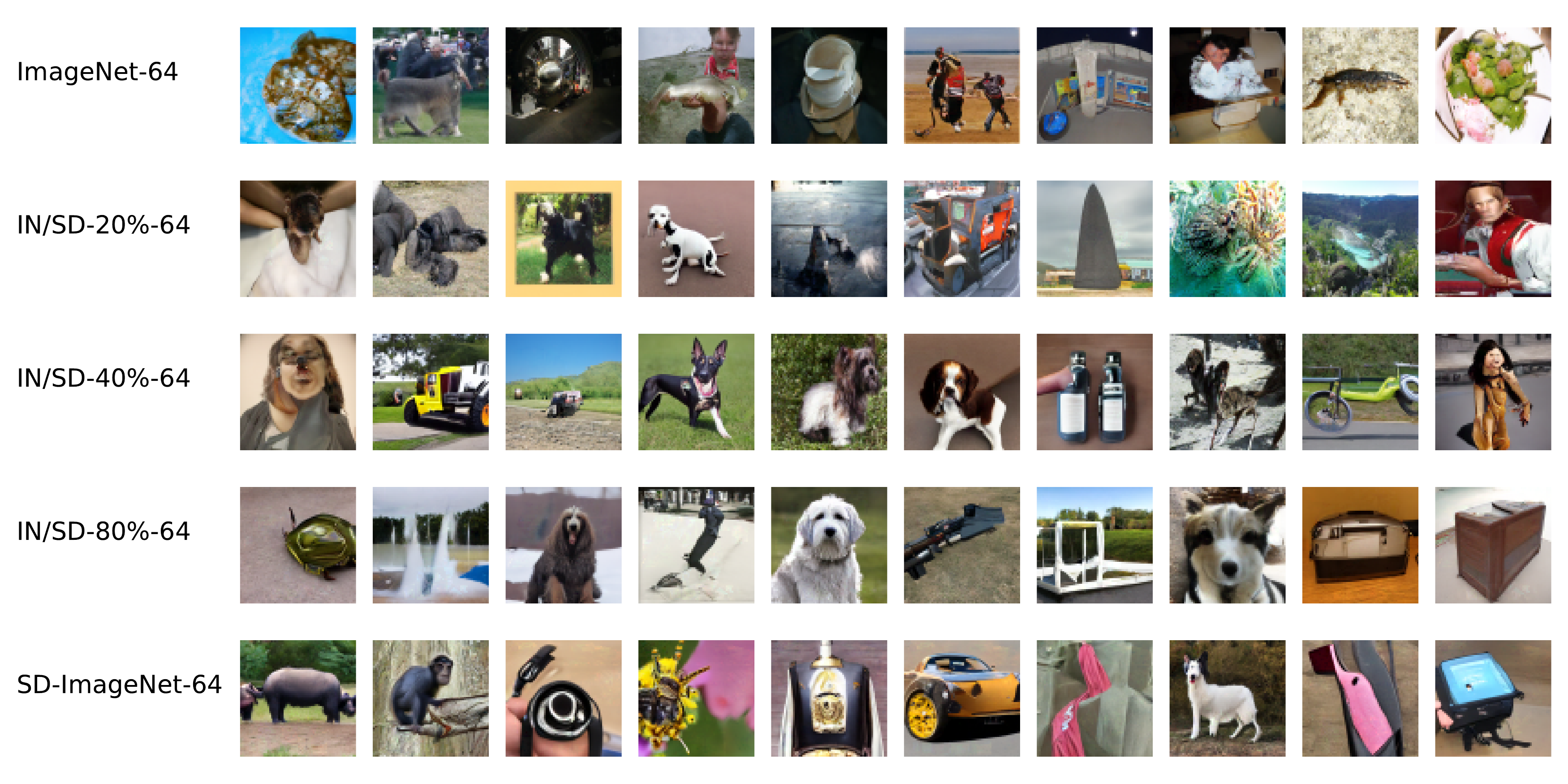}
    \caption{Randomly selected generated images by class-unconditional IDDPM~\cite{nichol21a} using 250 sampling steps.}
    \label{fig:generated_images}
\end{figure}

\Cref{tab:image_generation} reports the quality of unconditionally generated images in Fr\'echet Inception Distance (FID)~\cite{heusel2017gans}, improved precision and recall metrics of 5-nearest neighbors~\cite{kynkaanniemi2019improved} between Inception features of generated images and all validation data from ImageNet and WebVision.
Randomly sampled generated images are presented in \cref{fig:generated_images}.
The improved precision and recall metrics are computed by estimating the volumes of real and generated images in the embedded space with nearest neighbors~\cite{kynkaanniemi2019improved},
from which we can deduce how much two image distributions overlap with each other.
From \cref{tab:image_generation}, we see the trend that the precision and recall increases and decreases, respectively, as the ratio of generated images in training data increases.
To put it differently, the heavier contamination results in generated images that are more likely to be in the support of test images, while the test support coverage is worsened.
This indicates that generated images may concentrate on a smaller subset of the test support.

\begin{table}[t]
	\centering
	\resizebox{\linewidth}{!}{%
	\begin{tabular}{lccc}
	\toprule
	                & FID    $\downarrow$     &  Precision@5 $\uparrow$  &  Recall@5 $\uparrow$      \\
	\midrule
	ImageNet-64     & 14.9 / 15.6   & 0.665 / 0.679 & 0.644 / 0.653    \\
	\midrule
	IN/SD-20\%-64   & 12.6 / 12.7   & 0.699 / 0.708 & 0.621 / 0.634    \\
	IN/SD-40\%-64   & 11.0 / 10.8   & 0.730 / 0.739 & 0.585 / 0.608    \\
	IN/SD-80\%-64   & 12.5 / 11.3   & 0.795 / 0.802 & 0.490 / 0.512    \\
	SD-ImageNet-64  & 16.9 / 15.4   & 0.831 / 0.835 & 0.364 / 0.379    \\
	\bottomrule
	\end{tabular}
	}
 
	\caption{Image quality comparison on unconditional ImageNet-64 validation data and WebVision-64 validation data using Inception-V3 features shown in left and right of each cell, respectively.}
	\label{tab:image_generation}
\end{table}

\section{Analysis}\label{sec:analysis}

\subsection{Possible Cause of Degeneration}

Seeing the performance degeneration caused by contamination, we hypothesize that generated images have fewer modes than real ones.
To verify this idea, we measure the precision and recall of ImageNet training data and SD-ImageNet compared to ImageNet validation data, presented in \cref{tab:dataset_stats}.
As can be seen, the recall of SD-ImageNet images is significantly smaller than that of ImageNet images, indicating that SD-ImageNet images cannot cover the modes of ImageNet images, \ie, SD-ImageNet is less diverse than ImageNet.
The lack of diversity can also be observed visually, comparing images from ImageNet and SD-ImageNet in \cref{fig:sdimagenet_comparison} top and middle.
This observation can explain the performance degeneration in the main experiments, where contaminated datasets are concentrated in some modes, and thus models trained with them cannot generalize better.

\begin{table}[t]
    \centering
    \resizebox{0.8\linewidth}{!}{%
    \begin{tabular}{lcc}
    \toprule
    Dataset   &  Precision   & Recall \\
    \midrule
    ImageNet  &  0.757        & 0.791   \\
    Original SD-ImageNet & 0.836       & 0.344   \\
    Complex SD-ImageNet & 0.777 & 0.450 \\
    \bottomrule
    \end{tabular}
    }
    \caption{Precision and recall of the ImageNet training data and generated datasets compared with the ImageNet validation data using Inception-V3 features.}
    \label{tab:dataset_stats}
\end{table}

\subsection{Effects of Prompts}\label{sec:prompts}

\begin{figure}[h!]
    \centering
    \includegraphics[width=\linewidth]{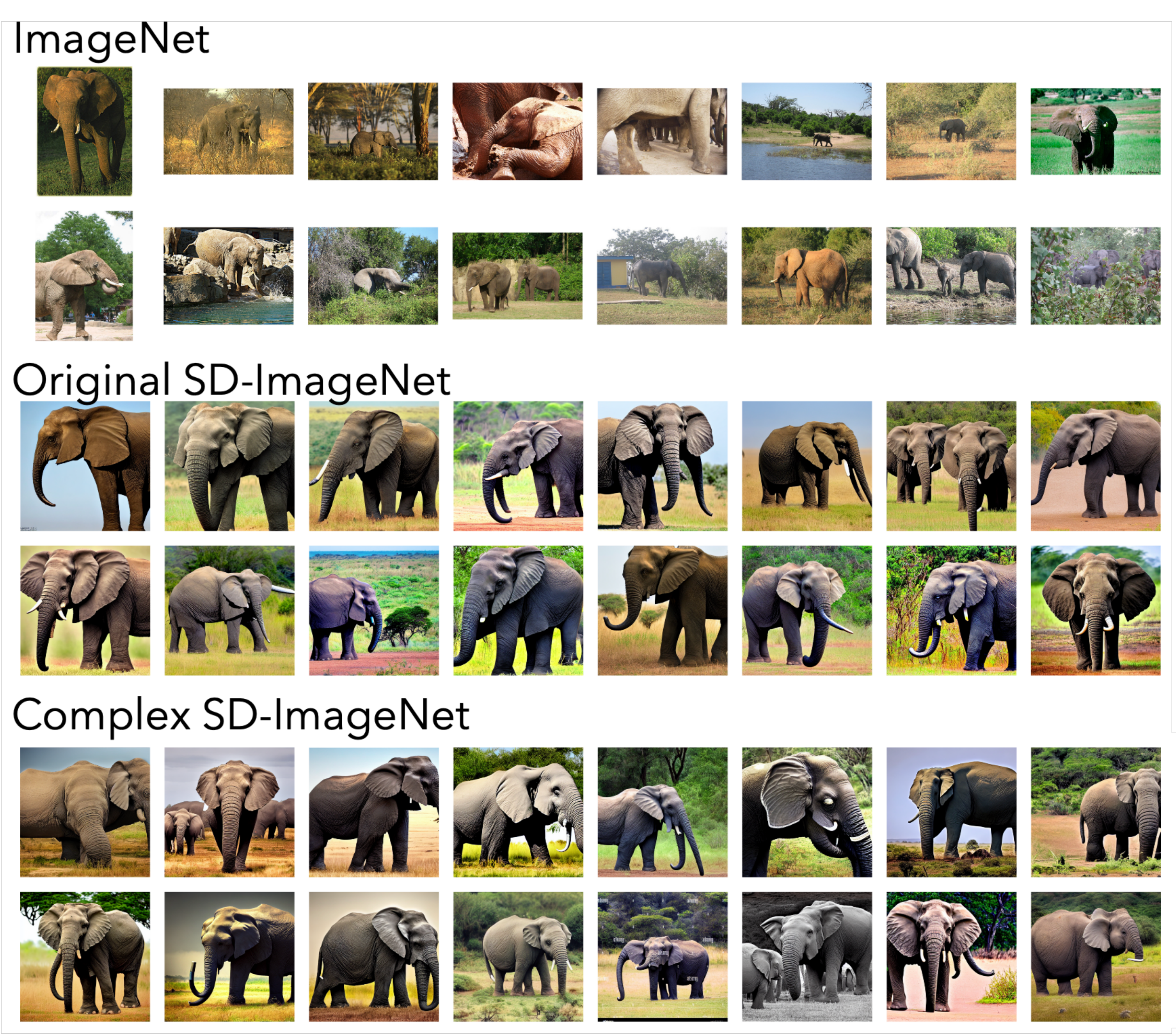}
    \caption{Randomly sampled images from the ``African elephant'' category of ImageNet, the original SD-ImageNet, and the complex SD-ImageNet. See \cref{fig:elephant_sd,fig:elephant_in,fig:elephant_sd2} in the Supplementary Material for more examples.}
    \label{fig:sdimagenet_comparison}
\end{figure}

As described in \cref{sec:dataset_generation}, the images of SD-ImageNet are generated from simple prompts like ``a photo of African elephant,'' which are unique to each class.
To verify the effects of prompt diversity, we create another SD-ImageNet, referred to as complex SD-ImageNet, with more complex prompts mimicking humans’, such as ``a monochrome image of African elephant taken with iPhone'' and ``HDR picture of grand piano outside.''
These prompts are programmatically generated from 200 to 1300 variations per class, and the details are explained in \cref{app:complex_prompts}.
\Cref{fig:sdimagenet_comparison} bottom illustrates samples from complex SD-ImageNet, \cref{tab:dataset_stats} bottom line shows quality of images, \cref{tab:sdimagenet-2} presents validation accuracy of ResNet-50 trained with the original and complex SD-ImageNets.
Although the complex SD-ImageNet's generation prompts are much more diverse than the original ones, the diversity of generated images is far less than real ones.
Consequently, the performance gain is marginal, indicating that our observation can be applied to predicting the problems caused by images conditionally generated from prompts composed by humans.

\begin{table}[h!]
    \centering
    \resizebox{0.95\linewidth}{!}{%
    \begin{tabular}{lccc}
         \toprule
         &  ResNet-50   &   Swin-S   &  ConvNeXt-T   \\
         \midrule
Original SD-IN &15.7        &  19.3      & 19.6       \\
Complex SD-IN     &18.6        &  21.9      & 23.1       \\
    \bottomrule
    \end{tabular}
    }
    \caption{Validation accuracy in classification of ResNet-50 trained with the original SD-ImageNet and complex SD-ImageNet generated from more complex prompts.}
    \label{tab:sdimagenet-2}
\end{table}

\subsection{Effects on Robustness}\label{sec:effect_on_representation}

In the main experiments of the classification task, we saw the performance only by accuracy on validation data.
To further investigate the effect of generated images on learned representation, we measured accuracy on other validation data; namely, ImageNet-A~\cite{imagenet-a} and ImageNet-R~\cite{imagenet-r}.
These datasets share the same categories with ImageNet, but are curated independently to measure robustness to out-of-distribution data.

\Cref{tab:robustness} summarizes the results, which generally indicate that generated images degenerate the robustness, except for IN/SD-20\% on ImageNet-A.
Contrarily, WebVision images may consistently enhance the robustness to out-of-distribution data on ImageNet-A and ImageNet-R, meaning that they are from a different distribution than ImageNet but diverse enough (compare, \eg, IN/WV-20\% to IN/WV-40\%).
These results further support the hypothesis that generated images have fewer modes than the real data, and thus, cause the downstream performance drop on test data and out-of-distribution data.

\begin{table}[t]
	\centering
	\resizebox{\linewidth}{!}{%
	\begin{tabular}{lccc}
		\toprule
		Source 		& ImageNet Val   & ImageNet-A   &  ImageNet-R  \\
		\cmidrule(ll){1-1}\cmidrule(rl){2-4}
		ImageNet    & 75.7           & 1.76                                           &  36.7                                \\
		\cmidrule(ll){1-1}\cmidrule(rl){2-4}
		IN/SD-20\%  & 74.5           & 2.12                                           &  35.7                             \\
		IN/SD-40\%  & 72.6           &  1.67                                           &  35.0                              \\
		IN/SD-80\%  & 65.3           &  1.61                                           &  30.0                                \\
		\cmidrule(ll){1-1}\cmidrule(rl){2-4}
		IN/WV-20\%  & 75.1           &  3.55                                          & 40.7                               \\
		IN/WV-40\%  & 73.9           & 3.77                                           & 41.5                               \\
		IN/WV-80\%  & 68.3            &5.00                                           & 40.0                              \\
		\bottomrule
	\end{tabular}
	}
	\caption{Robustness metrics in classification accuracy of ResNet-50 on the ImageNet validation set, ImageNet-A~\cite{imagenet-a}, and ImageNet-R~\cite{imagenet-r}. Different from IN/WVs, IN/SDs generally affect robustness to out-of-distribution data.}
	\label{tab:robustness}
\end{table}

\subsection{Comparison with Subsampled and added Data}\label{sec:data_augmentation}

In the main experiments, we compared the performances between networks trained with the contaminated datasets and with the \emph{full-size} clean datasets.
However, one may argue that the performance degradation results from the different amounts of real data in the training set.
In this section, we compare the performance of subsampled and added real datasets with contaminated datasets to disentangle the effect of the amount of clean data.

\Cref{tab:dataaugmentation} shows the validation accuracy of ResNet-50 trained with a 5\% subset of ImageNet and IN/SD-95\%, which fills 95\% of missing data by generated data.
Although IN/SD-95\% yields 7.4\% performance improvement over subsampled ImageNet, this is inferior to the gain by IN/WV-95\%.
We additionally measured the validation accuracy of ResNet-50 trained with full ImageNet (100\%) added with SD-ImageNet with different sizes.
\Cref{tab:add} shows the results of these experiments, indicating that SD-ImageNet images do not contribute to performance improvement even though the total dataset size increases.

In \cref{tab:dataaugmentation2}, test metrics of BLIP trained with 5\% and 20\% subsets of COCO and their corresponding CO/SDs are presented.
In this case, adding generated data affects negatively, even when only 5\% of real data are available.

Additionally, \Cref{tab:dataaugmentation3} compares the image quality of generated images by IDDPM trained with the IN/SD-80\% and a 20\% subset of ImageNet.
Aligned with the results in \cref{sec:experiments}, the recall metric diminishes with contaminated data, supporting the hypothesis that the modes of generated images are fewer than real ones.

These results emphasize the negative effects of generated data.
Additionally, the observations imply that using generated images for data augmentation needs careful consideration.
Such an idea has been studied in image classification using conditional GANs~\cite{Tran2017a,Antoniou2018b}, particularly in medical imaging~\cite{yi2019}, but also known to hinder the final performance in large-scale settings~\cite{Ravuri2019}.
Our results align with the latter that generated images are not always effective in data augmentation.

\begin{table}[t]
	\centering
    \resizebox{0.8\linewidth}{!}{%
	\begin{tabular}{ccc}
		\toprule
		ImageNet (5\%)      &    IN/SD-95\%    & IN/WV-95\%   \\
		\midrule
		44.7                 &   52.1          & 61.4         \\
		\bottomrule
	\end{tabular}
    }
	\caption{Validation accuracy of ResNet-50 trained with a 5\% subsampled ImageNet, IN/SD-95\%, and IN/WV-95\%.}
	\label{tab:dataaugmentation}
\end{table}

\begin{table}[t]
	\centering
    \resizebox{0.9\linewidth}{!}{%
	\begin{tabular}{lc}
		\toprule
                              & Accuracy \\
        \midrule
        ImageNet (100\%)      &   75.7       \\
        ImageNet (100\%) + SD-ImageNet (20\%)  &  75.7    \\
        ImageNet (100\%) + SD-ImageNet (40\%)  &  74.9    \\
        ImageNet (100\%) + SD-ImageNet (80\%)  &  75.0    \\
		\bottomrule
	\end{tabular}
    } 
	\caption{Validation accuracy of ResNet-50 when SD-ImageNet images are appended to the original ImageNet dataset.}
	\label{tab:add}
\end{table}

\begin{table}[t]
	\centering
    \resizebox{0.9\linewidth}{!}{%
	\begin{tabular}{lccc}
	\toprule
				& BLEU-4 $\uparrow$    & SPICE $\uparrow$     & CIDEr $\uparrow$    \\
	\midrule
	COCO (5\%) & 0.386     & 0.233      & 1.290     \\
	CO/SD-95\% & 0.362     & 0.226      & 1.220     \\
	\midrule
	COCO (20\%) & 0.385    & 0.235      & 1.305      \\
	CO/SD-80\%  & 0.377    & 0.233      & 1.279     \\
	\bottomrule
	\end{tabular}
    }
	\caption{Test metrics of BLIP trained with 5\% and 20\% COCO subsets and mixtures to complement the missing data. Higher values are better.}
	\label{tab:dataaugmentation2}
\end{table}

\begin{table}[]
    \centering
    \resizebox{\linewidth}{!}{%
    \begin{tabular}{lccc}
         \toprule
                        & FID $\downarrow$     &  Precision@5 $\uparrow$  &  Recall@5 $\uparrow$   \\
         \midrule
         ImageNet-64 (20\%)& 16.5    & 0.639           & 0.646        \\
         IN/SD-80\%-64     & 12.5    & 0.795           & 0.490         \\
         \bottomrule
    \end{tabular}
    }
    \caption{Image quality comparison of generated images of IDDPM trained with the IN/SD-80\% and a 20\% ImageNet subset.}
    \label{tab:dataaugmentation3}
\end{table}

\section{Discussion}\label{sec:discussion}

\subsection{Detection of Generated Images}\label{sec:detection}

\begin{figure}[t]
    \centering
    \includegraphics[width=0.9\linewidth]{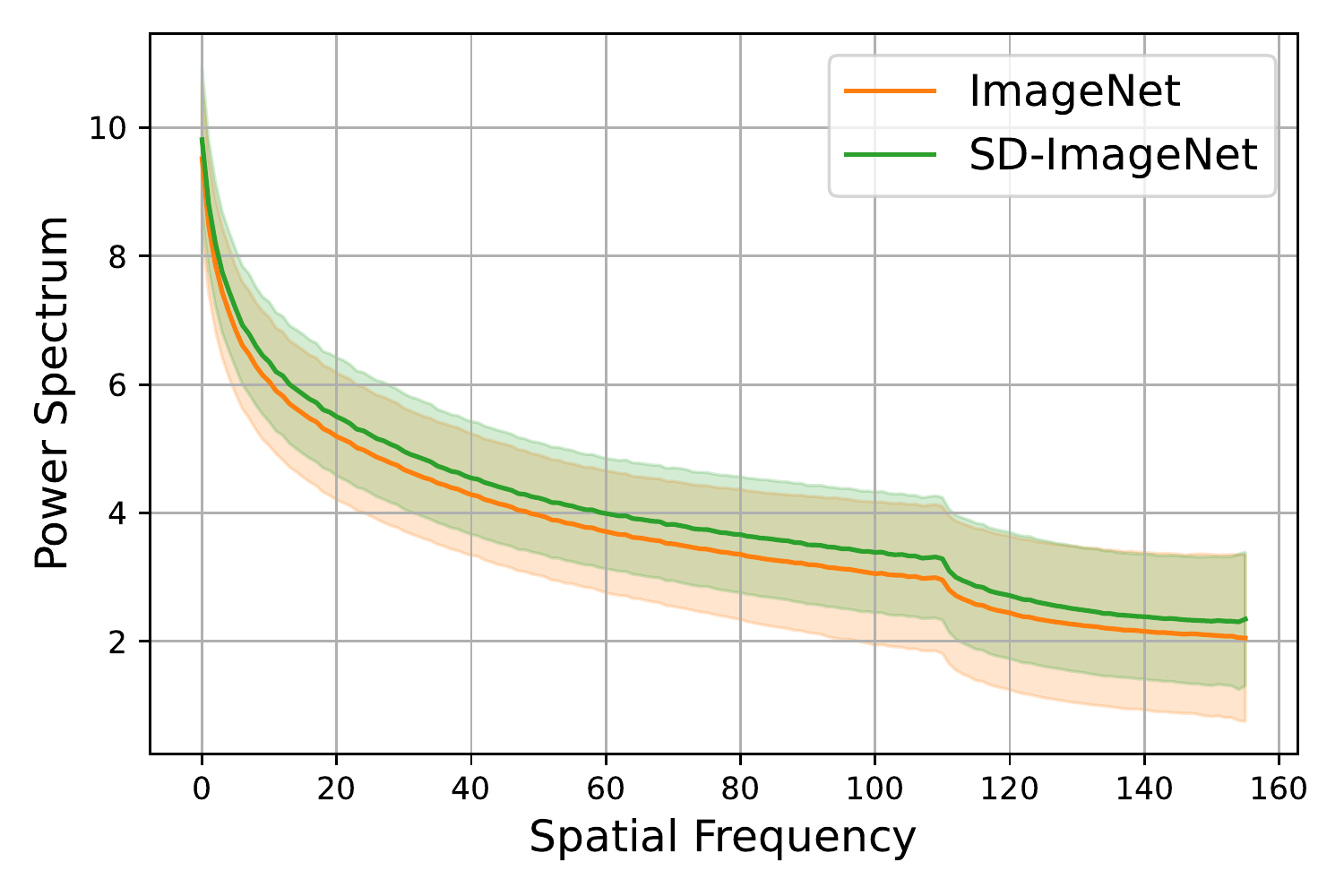}
    \caption{Average and standard deviation of spectra of $10^3$ images from ImageNet and SD-ImageNet.}
    \label{fig:1dfrequency}
\end{figure}

To avoid the negative effects by generated images, one may want to detect generated images easily.
For example, exploiting the differences between real and generated images in high-frequency spectra is a simple and convincing approach~\cite{dzanic2020}.
However, this discrepancy may be caused when an upsampling operation (to decode the original images from the low-dimensional latent representations) includes zero pixel insertion; otherwise, detecting generated images only by frequency spectra is difficult~\cite{chandrasegaran2021closer}.
Probably because StableDiffusion uses upsampling by nearest neighbor rather than zero pixel insertion, distinguishing real and generated images only from frequency information may be difficult.
\Cref{fig:1dfrequency} presents the power spectra of 1,000 images from ImageNet and SD-ImageNet, which are highly overlapped in all frequencies, which agrees with~\cite{chandrasegaran2021closer}.
Additionally, we trained a linear classifier and a multi-layer perceptron with ImageNet-pre-trained ResNet features to detect generated images.
When trained with $10^4$ images from both datasets, the classifiers achieved around 85\% test accuracy on 2,000 separated test images, which is still an unsatisfactory detection rate for a binary classification task and indicates the difficulty of detection of generated images.

\subsection{Self-supervised Learning as Remedy}

The experimental results so far indicate negative effects of generated images, and automatically filtering them out may be difficult (\cref{sec:detection}).
To alleviate the effect, one may be able to use self-supervised learning that aims to acquire useful representations without using explicit supervision.
To verify this idea, we adopt a self-supervised learning method called masked autoencoder (MAE)~\cite{he2022masked}.
We pre-train a Vision Transformer model, specifically ViT-B~\cite{dosovitskiy2021}, as MAE's encoder for 200 epochs with a mask ratio of 0.75.
\Cref{tab:ssl} presents validation accuracy after 90 epochs of linear probing that trains only the last classifier layer from the extracted features.
Even when the training dataset is fully generated images, the performance degeneration is limited, indicating that self-supervised learning may be a promising way to circumvent the negative effects of generated images.

\begin{table}[t]
	\centering
    \resizebox{0.5\linewidth}{!}{%
 	\begin{tabular}{lc}
	\toprule
	             & Accuracy  \\
	\midrule
	ImageNet     & 43.9           \\
	\midrule
	IN/SD-20\%   & 44.0           \\
	IN/SD-40\%   & 44.1           \\
	IN/SD-80\%   & 42.5           \\
	SD-ImageNet  & 38.8           \\
	\midrule
	IN/WD-20\%   & 43.3           \\
	IN/WD-40\%   & 43.3           \\
	IN/WD-80\%   & 40.9           \\
	WebVision    & 44.1           \\
	\bottomrule
	\end{tabular}
	}
	\caption{Validation accuracy of linear probing of MAE \cite{he2022masked} on the ImageNet validation set.}
	\label{tab:ssl}
\end{table}

\subsection{Limitations}

This paper has revealed the potential effects of generated images on datasets through various experiments.
Nevertheless, the discussion has some limitations.
Firstly, we could only use StableDiffusion trained with LAION-2B, because models and their pre-trained weights are publicly available, which is important to generate images without identifiable watermarks.
Different generative models and source datasets may lead to other conclusions, which are left for future work.

Another limitation is the types of created datasets and tasks of experiments.
Specifically, the datasets are created from synthetic prompts, and such a dataset generation scheme may be too simple to approximate possible data generation processes by users' prompts.
In addition, these datasets and tasks may often not be so complex that the insights of this paper would not cover some important aspects of other visual recognition tasks.
For example, the object counting task~\cite{chattopadhyay2017counting} on contaminated data may be challenging because generative models cannot always correctly handle numbers~\cite{saharia2022photorealistic}.
We leave a further in-depth analysis for future research.

Additionally, our experiments of classification and generation tasks were limited to the ``training from scratch'' paradigm.
Fine-tuning pre-trained models on carefully curated data might effectively circumvent the contamination issues.

\section{Conclusion}\label{sec:conclusion}

Recent generative models trained with billion-scale data enable to generate high-quality and high-fidelity images, and many users play with these models to share generated images on the web.
Observing such a trend, we questioned if such generated images affect the quality of future datasets collected images from the Internet.
To answer this question, we simulated contamination of generated images using a state-of-the-art generative model and conducted experiments on such data in various tasks, namely, image classification, image captioning, and image generation.
Throughout experiments, we found that generated images impact negatively on downstream performance, although its extent depends on the ratio of generated images and downstream tasks.
Additional analysis revealed that generated images degrade robustness to out-of-distribution data; application of generated images to data augmentation needs careful consideration; easy detection of generated images may not be applicable to up-to-date generative models; and self-supervised learning may be a promising remedy to the problem.

Based on these observations, we recommend that researchers to publish generative models carefully implement watermarks to enable the identification of generated images.
As we discussed in this paper, generated images have negative impacts on downstream performance, and their effect on new tasks is immeasurable; thus, publishers of generative models have the responsibility to avoid possible contamination.
One simple way to avoid this problem is to implement either identifiable or invisible watermarks, as some publishers have already done, \eg, \cite{dalle2,stablediffusion}, then dataset curators can easily identify and filter them out. 
We also suggest that researchers who develop image datasets collected from the Internet should filter out or mark generated images, which may affect final downstream performance, because adding generated images may degenerate performance as shown in \cref{sec:data_augmentation}.

Another important implication of this paper is further research on the detection methods of generated images, in parallel with the development of generative models.
As experimented in \cref{sec:detection}, generated images of the latest generative methods cannot be detected by simple methods that once had been effective.
Consequently, their development for filtering is essential for the soundness of future research.

\section*{Acknowledgement}

This work was supported by JST, ACT-X Grant Number JPMJAX210H, Japan.
We used computational resources of ``mdx: a platform for the data-driven future'' and RAIDEN (Riken AIp Deep learning ENvironment).
R.H. thanks Kai Katsumata at the University of Tokyo for his suggestions on image generation experiments.
We also appreciate constructive comments from the anonymous reviewers.

{\small
\bibliographystyle{ieee_fullname}
\bibliography{ref}
}

\clearpage
\appendix
\renewcommand\thefigure{\thesection.\arabic{figure}}
\setcounter{figure}{0}

\section*{Supplementary Material of ``Will Large-scale Generative Models Corrupt Future Datasets?''}

This supplemental material describes experimental settings (\cref{app:config}) and example images from ImageNet and SD-ImageNet (\cref{app:example_images}).

\section{Detailed Experimental Configurations}\label{app:config}

This section describes the detailed experimental settings and configurations.

\subsection{Dataset Creation}

We generated images using the StableDiffusion model\footnote{\url{https://github.com/CompVis/stable-diffusion}} and its accompanying pre-trained weight (\texttt{sd-v1-1.ckpt}) on eight NVIDIA A100 GPUs.
Each image of the datasets was sampled by 50 steps of the PLMS sampler with an unconditional guidance scale of 7.5, which is identical to the setting of its web application.\footnote{\url{https://huggingface.co/spaces/stabilityai/stable-diffusion/blob/main/app.py}}

\subsection{Image Classification}

We trained ResNet-50 and Swin-S models following \texttt{torchvision}'s training protocol\footnote{\url{https://github.com/pytorch/vision/tree/v0.13.1/references/classification}} on eight NVIDIA A100 GPUs.
The results of Swin-S were calculated using parameters with exponential moving average.

\subsection{Image Captioning}

We fine-tuned the captioner and filter modules of the BLIP model following \texttt{LAVIS}'s training script\footnote{\url{https://github.com/salesforce/LAVIS/blob/v0.1.0/run_scripts/blip/train/train_caption_coco.sh}} on two NVIDIA A100 GPUs.

\subsection{Image Generation}

We trained and generated images from IDDPM following the official instructions for the ImageNet-64 dataset\footnote{\url{https://github.com/openai/improved-diffusion/tree/main}} on eight NVIDIA V-100 GPUs.
The model was trained for $1.8\times 10^6$ iterations using the $L_\text{hybrid}$ objective with a batch size of 512.
We generated 50,000 images with 250 sampling steps from EMA models.
The computation of metrics is based on \url{https://github.com/NVlabs/stylegan2-ada-pytorch/tree/main/metrics}.

\subsection{Complex Prompts}\label{app:complex_prompts}

We synthetically generated complex prompts for each ImageNet category using the following script.

\begin{lstlisting}[language=Python,basicstyle=\footnotesize]
import numpy as np

def generate_prompt(category_names: list[str]) -> str:
    name = np.random.choice(category_names, 1)
    _0 = ['', 'high quality', 'low quality', 
          'monochrome', 'blured', 'atmospheric', 
          'rendered', 'zoomed', 'wide-angle',
          'hdr', 'high resolution']
    _1 = ['photo', 'picture', 'realistic photo', 
          'image']
    _2 = ['', 'taken with iPhone', 'inside', 
          'outside', 'without background']

    _0 = np.random.choice(_0, None)
    _1 = np.random.choice(_1, None)
    _2 = np.random.choice(_2, None)
    return f"{_0} {_1} of {name} {_2}".strip()
\end{lstlisting}

The words modifying the prompts are selected to mimic human prompts and be applicable to all classes in ImageNet.
Because many classes have multiple category names, \eg, ``African elephant'' and  ``Loxodonta africana'' for the African elephant class, this script can generate a variety of prompts, namely from 200 to 1300 different prompts per class.

\subsection{Self-supervised Learning}

We pre-trained MAE following the official implementation\footnote{\url{https://github.com/facebookresearch/mae/tree/main}} on eight NVIDIA A-100 GPUs and fine-tuned the last layer of its encoder on 16 NVIDIA V-100 GPUs.
Pre-training was for 200 epochs with a 40-epoch warmup with a batch size of 4096, using gradient accumulation once in every two iterations.
Fine-tuning was for 90 epochs using the LARS optimizer with a batch size of 16,384.

\subsection{Detection of Generated Images}

For the experiments in Section 6.1, we first extracted the features of ImageNet-pre-trained ResNet-50 on 12,000 images from ImageNet and SD-ImageNet.
Each feature vector has a dimension of 2,048.
Then, we trained a linear classifier and a two-layer MLP with a hidden size of 1,024 with a ReLU activation to classify them using 10,000 feature vectors for 5,000 iterations using the Adam optimizer with a batch size of 128.
Their performances were evaluated on the other 2,000 test vectors.
The linear classifier and the MLP achieve 83\% and 86\% accuracy, respectively.

\section{Additional Results}\label{app:example_images}

\subsection{Comparison of Real and Generated Images}

\begin{figure}[h!]
    \centering
    \includegraphics[width=0.9\linewidth]{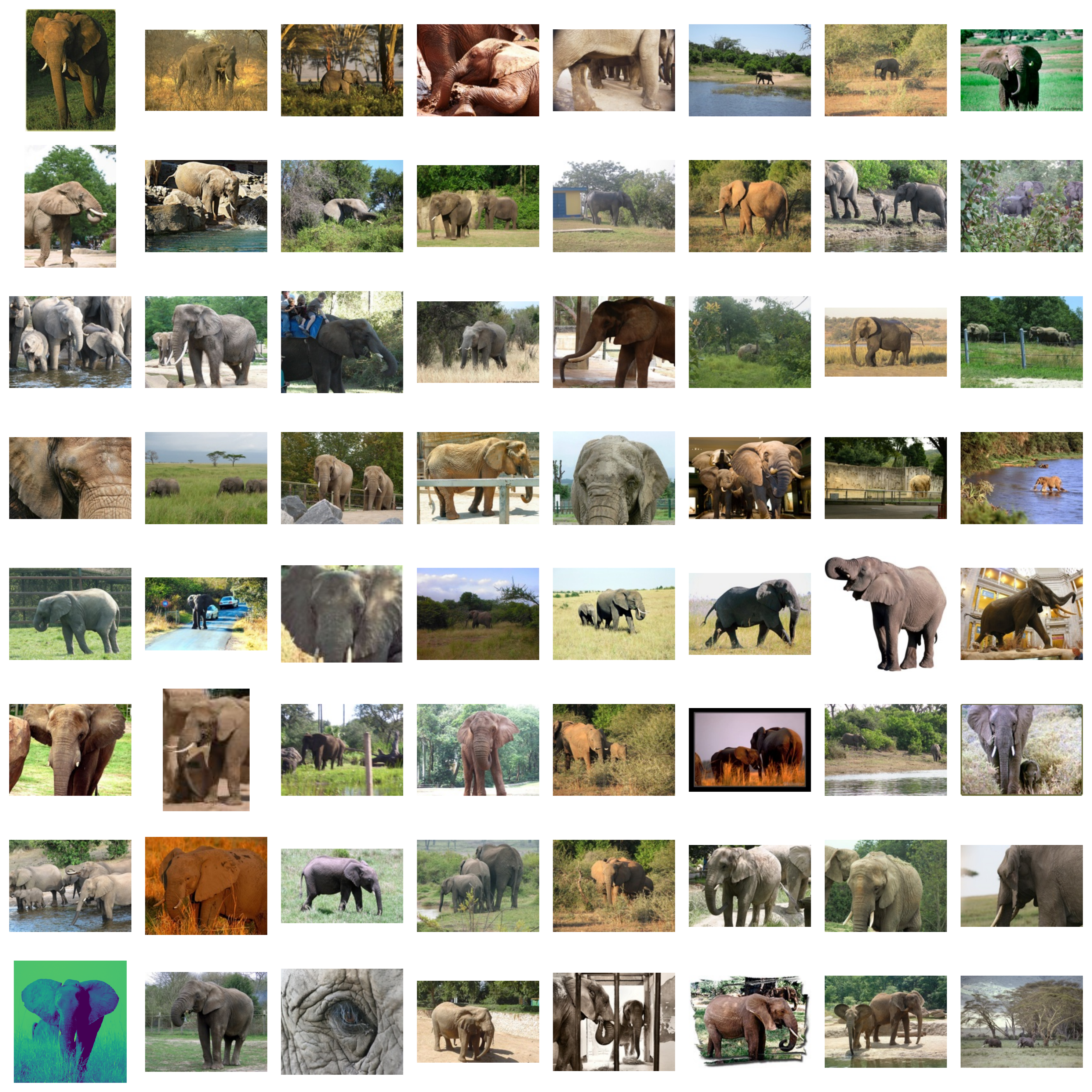}
    \caption{Real images of African elephants from ImageNet.}
    \label{fig:elephant_in}
\end{figure}

\begin{figure}[h!]
    \centering
    \includegraphics[width=0.9\linewidth]{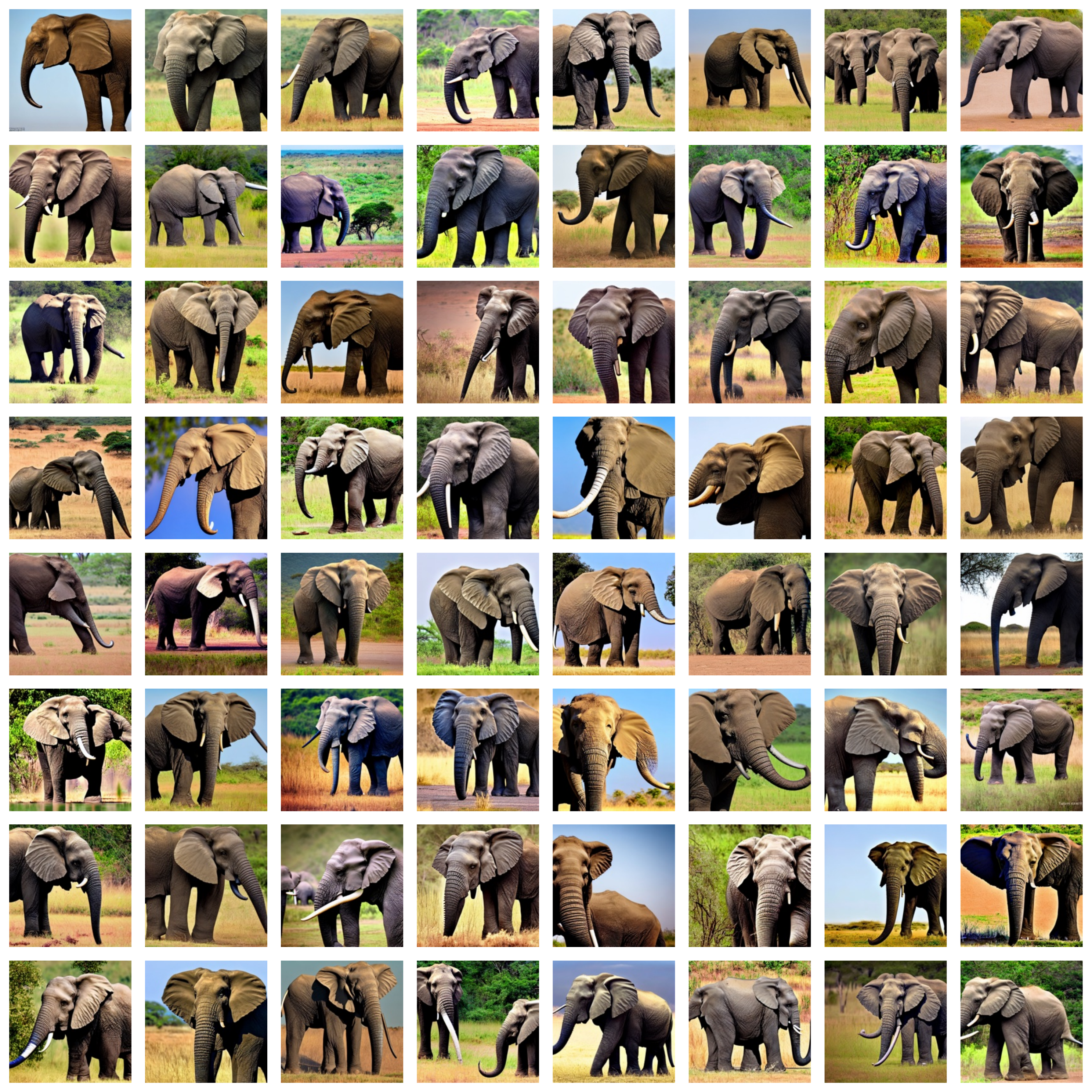}
    \caption{Generated images of African elephants from the original SD-ImageNet.}
    \label{fig:elephant_sd}
\end{figure}

\begin{figure}[h!]
    \centering
    \includegraphics[width=0.9\linewidth]{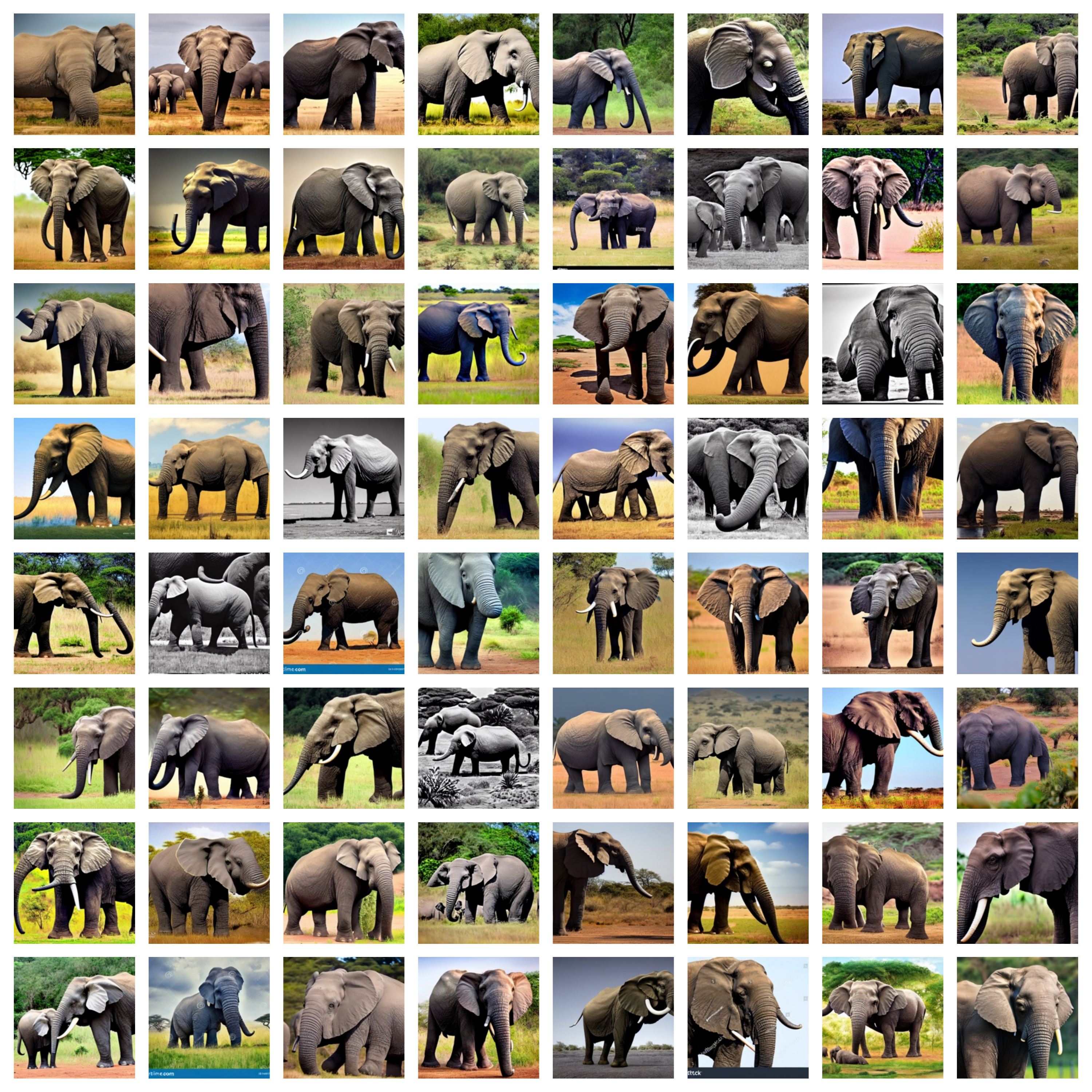}
    \caption{Generated images of African elephants from the complex SD-ImageNet.}
    \label{fig:elephant_sd2}
\end{figure}

In Section 4.4, we hypothesized that generated images have fewer modes than real ones, which causes the performance degeneration.
Comparing randomly selected images from ImageNet and SD-ImageNet in \cref{fig:elephant_sd,fig:elephant_in,fig:elephant_sd2} visually supports this hypothesis.

\subsection{Examples of \texttt{titi}}

\begin{figure}[h!]
    \centering
    \includegraphics[width=0.9\linewidth]{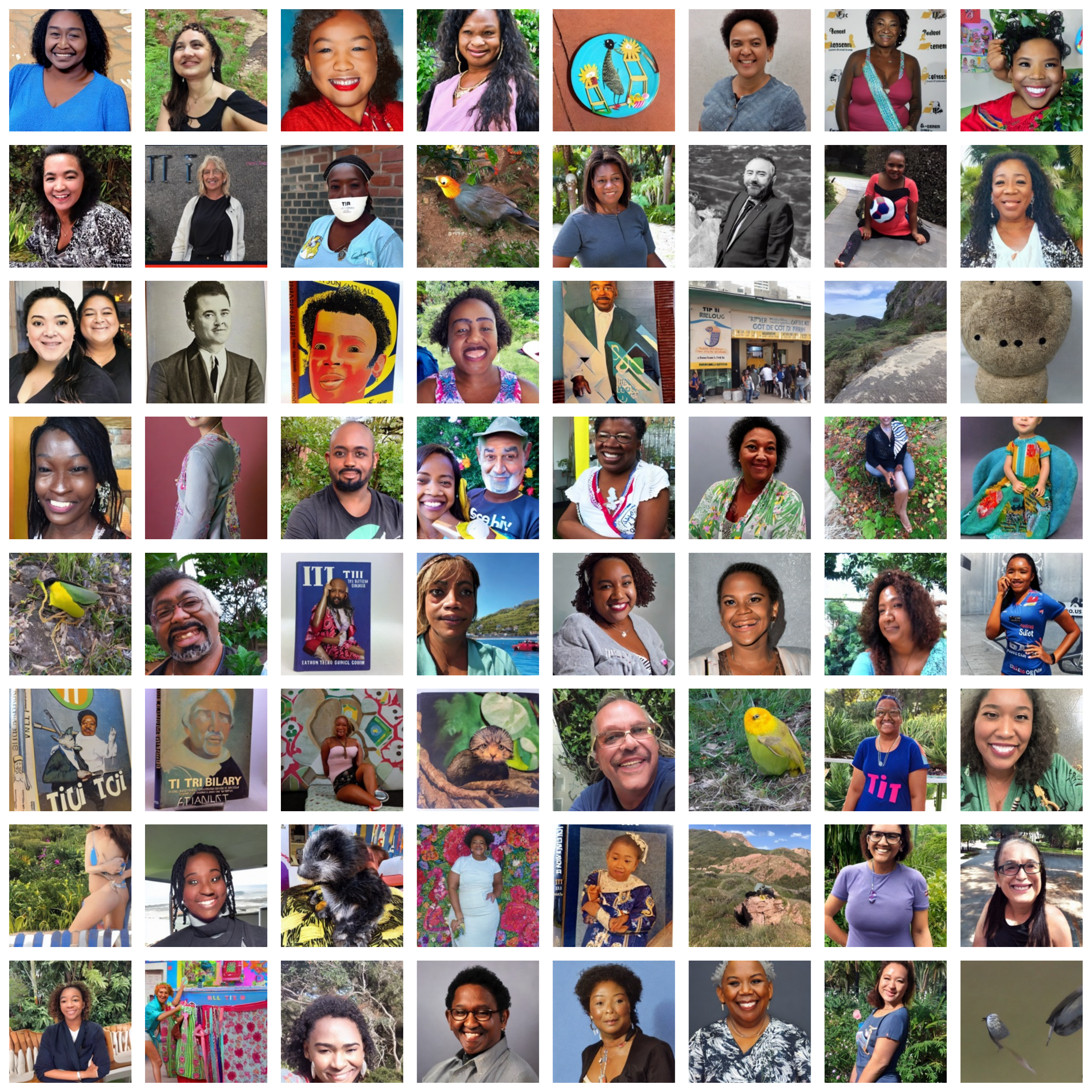}
    \caption{Generated images of the \texttt{titi} category from SD-ImageNet.}
    \label{fig:titi}
\end{figure}

In Section 4.1, we argued that some categories were semantically diverse because the ambiguity of category names.
\Cref{fig:titi} presents randomly selected images from the \texttt{titi} category.
Although ImageNet intended this class to mean a New World monkey, the generated images are mostly photos of humans, because ``titi'' is also a name of people.

\end{document}